# Quantifying the Influences on Probabilistic Wind Power Forecasts

Jens Schreiber and Bernhard Sick
Intelligent Embedded Systems Group, University of Kassel
Kassel, Germany
e-mail: {jens.schreiber, bsick}@uni-kassel.de

*Abstract*— In recent years, probabilistic forecasts techniques were proposed in research as well as in applications to integrate volatile renewable energy resources into the electrical grid. These techniques allow decision makers to take the uncertainty of the prediction into account and, therefore, to devise optimal decisions, e.g., related to costs and risks in the electrical grid. However, it was yet not studied how the input, such as numerical weather predictions, affects the model output of forecasting models in detail. Therefore, we examine the potential influences with techniques from the field of sensitivity analysis on three different black-box models to obtain insights into differences and similarities of these probabilistic models. The analysis shows a considerable number of potential influences in those models depending on, e.g., the predicted probability and the type of model. These effects motivate the need to take various influences into account when models are tested, analyzed, or compared. Nevertheless, results of the sensitivity analysis will allow us to select a model with advantages in the practical application.

## I. INTRODUCTION

In past years, renewable energy resources have become a fundamental part of the electrical power supply in many countries. In Germany, e.g., renewable resources contribute up to 29 percent to the energy mix [1]. A conventional approach to integrate those volatile resources into current grids is to use *numerical weather prediction (NWP)* data and historical power data as training data for deterministic machine learning models that generate day-ahead forecasts. Here, we call these models deterministic because their output is a point estimate of the future power production.

Recently, probabilistic forecasts can be seen in research as well as in applications [2]. Probabilistic methods forecast distribution of wind power rather than delivering a point forecast of the expected power. *Probabilistic forecasts (PF)* have, in comparison to deterministic approaches, the advantage that power plant operators, grid operators, and energy traders can examine the risk of their day-ahead forecasts. These probabilistic forecasts help decision makers to devise optimal decisions considering the uncertainty of the prediction, mainly caused by the NWP [3].

However, even though many studies investigate influences and interactions of input features on the output of deterministic forecasts, e.g., [4]–[9], there has been little interest in conducting studies that are designed to evaluate influences in probabilistic forecasts. This article examines influences of features on the output of three different probabilistic wind power forecasting models with a focus on NWP data and technical features (e.g., maximum diameter and maximum power generation of a wind turbine) as input (which is typical for day-ahead forecasts). Therefore, we investigate these potential influences with methods from the field of sensitivity analysis regarding the three models as black-boxes.

This analysis allows us to point out a considerable number of potential influences in probabilistic forecasts. Influences are, e.g., the forecasting model, the predicted quantile, as well as selected input features. That is, the causes of forecasts errors might be various. Therefore, we need to take several influences into account when models are tested, analyzed, and compared or simultaneously used within an application. In the end, results of the analysis lead us to a gray-box model, where we have partial knowledge about the internal behavior of the forecasting model. This knowledge will allow us to select a model with advantages in the practical application.

The remainder of this article is structured as follows. First, we give details on related work in Section II. We continue with explanations about probabilistic forecasts and sensitivity analysis in Section III. Then the experiment is presented in Section IV. Section V summarizes and concludes our work.

## II. RELATED WORK

Determining influences in forecasting models are an essential research topic in renewable resources. On the one hand, findings can help to improve the forecasting accuracy. On the other hand, insights help decision makers to understand their forecasting model. By understanding the relationship between input and output, it is, e.g., possible to determine potential causes of a forecasting error.

The most common methodology to analyze the influences on the output of the deterministic forecasts model is sensitivity analysis. Sensitivity analysis is, e.g., used in deterministic forecasts for investment calculation [10]. In [4], the authors use sensitivity analysis to proof the feasibility to integrate wind power generators into the Brazilian electricity market. Another application of sensitivity analysis shows the profitability and vulnerability of renewable energy resources [7]. In [5], sensitivity analysis is used to examine influences of features for the placement of floating offshore wind farms for future investments. It is also possible to develop a framework to examine and evaluate features that allow the assessment of potential locations for new wind or solar parks as shown in [6]. The authors of [8] use sensitivity analysis and swarm optimization to obtain the best placement of wind turbines in wind parks [8]. In [9], sensitivity analysis is used to show that weather data has a higher impact than technical features in wind power time series modeling. However, to our knowledge there is no article that attempts to apply sensitivity analysis to the analysis of models used for probabilistic (wind power) forecasting.

### III. METHODOLOGY

In this section, we give a brief introduction into probabilistic forecasts and sensitivity analysis.

#### A. Probabilistic Wind Power Forecasts

In this section, we focus on the background of probabilistic models, necessary to apply sensitivity analysis later.

Probabilistic methods forecast distribution of wind power rather than delivering a point forecast of the expected power. By predicting, e.g., a *cumulative distribution function (CDF)*, we obtain additional information about the uncertainty of our model. Uncertainties arise, e.g., due to the influence of the uncertain *NWP* based model inputs.

Typically, to predict those distributions one either uses parametric or non-parametric methods. Non-parametric methods have the benefit that they do not require assumptions about the mathematical form of the distribution of wind power. Three non-parametric models will be used in this article to forecast the wind power distribution. In particular, the techniques *monotone quantile regression neural network (MQRNN), support vector regression (SVR)* and, *gradient boosting regression tree (GBRT)* are used. For details, please refer to [11]–[14].

Each of the three approaches is considered as a black-box method to forecast an *empirical cumulative distribution function (ECDF)*. In contrast to a CDF, the ECDF is co posed of several quantiles and it is linearly interpolated between the respective quantile estimates.

#### B. Sensitivity Analysis

*Sensitivity analysis (SA)* typically decomposes the uncertainty of a single (deterministic) output to different sources of uncertainty in the input [10]. However, we are interested in applying SA to the predicted ECDF from three black-box models and not to a deterministic output (i.e., point estimate). Therefore, for each predicted quantile SA needs to be applied. Each quantile can be regarded as a single predicted output of the ECDF. This utilization allows us to evaluate each quantile individually without adapting the method itself. In this sense, by applying SA to each quantile, we examine how individual quantiles are influenced differently by the input. Further on, by comparing the results for different quantiles, we can study how these influences are related. For better understanding, we will limit the following explanations to a single output.

One standard SA method to examine influences and interactions of inputs is the *variance based decomposition (VBD)* [10]. VBD uses Monte Carlo simulation to decompose the variance of the output $V(Y)$ w.r.t. the input. VBD, therefore, distinguishes between a *first-* and a *total-order sensitivity index* given by the following formulas:

$$S_i = \frac{V[E(Y|X_i)]}{V(Y)}, \quad (1)$$

$$S_{T_i} = \frac{V[E(Y|X_{\sim i})]}{V(Y)}, \quad (2)$$

where $S_i$ is the first-order sensitivity index and $S_{T_i}$ the total-order sensitivity index. $S_i$ can be interpreted as the degree of influence of feature *i* on the output. The first-order sensitivity index is the relation between the variance, $V[E(Y|X_i)]$, in the output that is explained by feature *i* alone and the variance in the output.

$S_{T_i}$ is the first-order index of feature *i* plus all higher-order interactions of this feature, e.g., the effect of wind speed alone plus the effect together with air pressure. More precise, in the field of SA, the interaction is defined as the effect that cannot be explained by a single effect alone [10]. $X_{\sim i}$ indicates all possible combination of other features with feature *i*. For further details refer to [10], [15].

Note that a notable difference between $S_i$ and $S_{T_i}$ indicates strong interactions between features that affect the output together.

### IV. EXPERIMENTAL RESULTS

This section investigates the influences of three probabilistic wind power forecasting models, their input features, and the predicted quantiles with sensitivity analysis in three use cases. The SA of one use case along with one forecasting model is called *scenario* for convenience. First, we give our definition of the three use cases. Second, we explain the manual feature engineering and the preselection of features. This set of selected features allows us to get proper evaluations results of all three black-box models as detailed in the next section. The evaluation results are critical to assure that the forecasts of the wind power distribution are reasonable. Finally, we apply SA and get valid results that are summarized and discussed in the final sections.

#### A. Definition of Use cases

To examine the relation between input and output in probabilistic wind power generation, we evaluate three different use cases. The evaluation of the different use cases helps to distinguish different influences of the terrain and the relation as modeled by the respective PF model. Similar to other studies of wind power forecast models, see, e.g. [3], we define the following use cases:

- Wind parks located in *non-complex terrain (NCT)* between 200 and 1500 meters above sea level (e.g., farmland).

- Wind parks located in *complex terrain(CT)* between 200 and 1500 meters above sea level (e.g., forest).
- O*ffshore (OS)* wind parks located on the ocean.

Thee definitions allow us to evaluate 11 NCT, eight CT and four OS wind parks located in Germany from our data.

*B. Feature Engineering and Pre-Selection of Input Data*

This sections details how we manually engineered features and use feature selection to devise an optimal set of features. This set of features allows us to have reasonable forecast results in each scenario (summarized in the next section). All wind parks of the use cases have a resolution of one hour from 2015-01-01 to 2016-12-31. Initially, the data has the following properties:

- NWP features: *Air pressure (AP), humidity (H), wind direction zonal (WDZ100m) at 100m, wind direction meridional (WDM100m)* at 100m, wind speed (WS10m) at 10m, and *wind speed (WS100m)* at 100m above ground.
- Meta features: Maximum power generation, maximum diameter, maximum hub height, and elevation of the wind park.
- Normalization: Min-max normalization is applied to all NWP features initially. Respectively, derived features are normalized as well. The generated power is normalized with the maximum power generation.
- From the NWP features we derive *variability (V)* features. Variability is defined with $V = |F(t+k) - F(t)|$. This derived feature indicates the mean amount of change of a feature $F$ within a time horizon $k$ [16], e.g. of the *previous (P)* hour. For $k$ we used *hour (HR), day (D), week (W), month (M),* and *year (Y)*. Note that, e.g., the variability of wind speed at 100 meters in the previous hour is abbreviated with *VWS100mPHR*.
- We add features for the day, the week, the month of the year, and *hours since last model run (HSMR)* of the NWP to cover, e.g., trends and seasonal effects.

Afterward, the data is split into 80% training (January 2015 to July 2016) and 20% test data-set (August 2016 to December 2016). This data setup is used to pre-select features for each use case on the training data.

We use a combination of *sequential forward selection (SFS)* and two filters (Minimum Redundancy Maximum Relevance and Fisher Score [17]). SFS selects the ten most prominent features with GBRT and so-called *continuous ranked probability score (CRPS)* as selection criterion [18]–[20]. For the sake of this article, one can imagine the CRPS as the mean absolute error from deterministic forecasts. In parallel, we require that results of the SFS to be in the top ten ranked features from one of the filters.

This selection strategy improves the CRPS by a minimum of 1.6 percent and a maximum of 12 percent compared to filter or wrapper feature selection methods. However, the detailed evaluation is out of scope for this article.

*C. Probabilisitic Models Analyzed by Sensitivity Analysis*

The configuration of the models can be summarized as follows:

- The quantiles 0.1, 0.2, 0.3, 0.4, 0.5, 0.6, 0.7, 0.8, and 0.9 are used and analyzed in all scenarios.
- Input features for the scenario depends on the selected feature for the respective use case.
- All three models are trained and optimized with standard parameters on the train and validation data sets.

For the SA detailed evaluation of the error score is not relevant. However, it may be relevant to know that they achieved reasonable results. GBRT achieved the smallest average CRPS (0.061) and standard deviation (0.019). MQRNN has the second smallest mean (0.07) and standard deviation of the mean (0.025). SVR has the largest standard deviation (0.027) and average CRPS (0.073).

|  | GBRT | | | | | | MQRNN | | | | | | SVR | | | | | |
|---|---|---|---|---|---|---|---|---|---|---|---|---|---|---|---|---|---|---|
|  | OS | | NCT | | CT | | OS | | NCT | | CT | | OS | | NCT | | CT | |
| 1 | - | - | 0.02 | 0.04 | 0.02 | 0.04 | - | - | 0.02 | 0.23 | 0.02 | 0.14 | - | - | 0.01 | 0.23 | 0.03 | 0.26 |
| 2 | 0.00 | 0.01 | 0.00 | 0.01 | 0.00 | 0.00 | 0.00 | 0.08 | 0.00 | 0.07 | 0.00 | 0.05 | 0.03 | 0.31 | 0.03 | 0.30 | 0.02 | 0.28 |
| 3 | - | - | 0.01 | 0.03 | 0.00 | 0.01 | - | - | 0.01 | 0.14 | 0.00 | 0.10 | - | - | 0.03 | 0.30 | 0.02 | 0.25 |
| 4 | - | - | - | - | 0.01 | 0.03 | - | - | - | - | 0.01 | 0.13 | - | - | - | - | 0.02 | 0.24 |
| 5 | 0.01 | 0.03 | 0.02 | 0.03 | - | - | 0.04 | 0.41 | 0.01 | 0.21 | - | - | 0.02 | 0.22 | 0.03 | 0.22 | - | - |
| 6 | 0.01 | 0.03 | - | - | - | - | 0.03 | 0.38 | - | - | - | - | 0.04 | 0.25 | - | - | - | - |
| 7 | - | - | 0.00 | 0.01 | 0.00 | 0.01 | - | - | 0.01 | 0.12 | 0.00 | 0.08 | - | - | 0.06 | 0.33 | 0.05 | 0.29 |
| 8 | - | - | 0.00 | 0.01 | 0.01 | 0.03 | - | - | 0.01 | 0.13 | 0.01 | 0.08 | - | - | 0.05 | 0.35 | 0.05 | 0.31 |
| 9 | 0.68 | 0.82 | 0.82 | 0.89 | 0.86 | 0.91 | 0.32 | 0.68 | 0.49 | 0.73 | 0.69 | 0.84 | 0.23 | 0.53 | 0.17 | 0.47 | 0.18 | 0.49 |
| 10 | 0.09 | 0.19 | 0.04 | 0.07 | 0.02 | 0.03 | 0.07 | 0.41 | 0.06 | 0.27 | 0.02 | 0.16 | 0.10 | 0.39 | 0.08 | 0.35 | 0.10 | 0.37 |
|  | $S_i$ | $S_{T_i}$ | $S_i$ | $S_{T_i}$ | $S_i$ | $S_{T_i}$ | $S_i$ | $S_{T_i}$ | $S_i$ | $S_{T_i}$ | $S_i$ | $S_{T_i}$ | $S_i$ | $S_{T_i}$ | $S_i$ | $S_{T_i}$ | $S_i$ | $S_{T_i}$ |

**Table 1:** Results of the averaged first- and total-order sensitivity index $S_i$ and $S_T$ for the selected features for each scenario are shown. For each scenario, the sensitivity index is color-coded with green being the highest and red the smallest value. The relevant features are enumerated on the left in the following order AP (1), HSMR (2), H (3), T (4), VWS100mPHR (5), VWS10mPHR (6), WDM100m (7), WDZ100m (8), WS100m (9), and WS10m (10). The dashed line indicates that the specific feature is not selected in the scenario.

## D. Sensitivity Analysis

The following section summarizes results of the SA. In each use case, the same features are used; those features are selected beforehand by the feature selection strategy as described above to take the specifics of the terrain into account. The models trained on this data (for each scenario) are evaluated here with SA.

For all scenarios, the VBD sensitivity analysis is applied to each quantile separately. The Monte Carlo simulation uses 10.000 sample points. Within a use case, the same sample points are used for all models and quantiles.

Figure 1 shows three examples of the *average quantile first-order index*. By averaging the first-order index (of the same quantile) for all wind parks in a scenario, we obtain the average influence of features for a quantile in this scenario. The three machine learning examples are representative concerning the relation of quantiles. It shows, e.g., how an increase of $S_i$ for WS100m for different quantiles is related to a decrease of $S_i$ for WS10m (see, e.g., Figure 1a).

Table 1 shows the average $S_i$ and $S_{T_i}$ across all quantiles and wind parks within a scenario. In contrast to the average of Figure 1, the first-order or total-order is additionally averaged for all quantiles. This allows obtaining a single value that is representative of the scenario. The difference between these average $S_i$ and $S_{T_i}$, is used to examine the amount of interaction of features given by an increase between the first and total-order index. As mentioned before, in the field of SA interactions are defined as the effect on the output which cannot be explained by a single feature. The higher the increase from the average $S_i$ to the $S_{T_i}$, the larger is the amount of interaction of the feature with other features that affect the output together.

**Observations related to features:**
- As expected, WS100m is the essential feature in all evaluated scenarios. Similar, in most cases WS10m has the second largest total-order index.
- If selected, AP and VWS100mPHR have about the third largest total-order sensitivity index for all scenarios.

**Observations related to PF models:**
- The average total-order sensitivity index of Table 1 for GBRT is small for most features except WS100m and WS10m. In contrast, the values for SVR and MQRNN are more spread among all features. The results for GBRT are potentially caused by the ensemble of weak predictors. The combination of those predictors allows the GBRT to achieve the best forecast results without substantial interaction with other features compared to SVR and MQRNN. For SVR one can assume that the worst evaluation result is related to the extensive amount interactions between all features. These interactions potentially cause too much variability in the output and yield to the worst evaluations results. MQRNN is somewhat in the middle of the evaluation score and the interactions of features. Potentially, the internal data transformation by the MQRNN allows the model to capture interactions of features and keep the variance in the output to a minimum at the same time. Respectively, this data transformation allows MQRNN to achieve better results than the SVR.
- SA shows that influences of features are about the same for identical types of PF methods across different use cases. This observation indicates that the internal structures of the black-box models are more affected by the underlying model than by the data.

**Observations related to quantiles:**
- For similar PF model, the relation of the first and total-order index for different quantiles are about the same for all use cases. E.g., the initial decrease between the 0.1 and 0.4 quantile in Figure 1c is similar in all SVR scenarios. This observation further motivates that the relationship between input and output is largely depending on the underlying PF model and not the specifics of the input data.
- SA shows that for SVR and GBRT the amount of influence is dependent on the quantile of the predicted ECDF.

## E. Discussion

Interestingly, SA shows that influences and interactions of features mostly depend on the underlying model. This result partly surprises, because in existing studies of the uncertainty, the forecast error is seen as being largely dependent on the terrain, see, e.g. [3]. Therefore, studies of the uncertainty, as in [21], would probably benefit substantially from SA to understand further causes of the error. On the other hand, it might also be related use case specific selection of the input features.

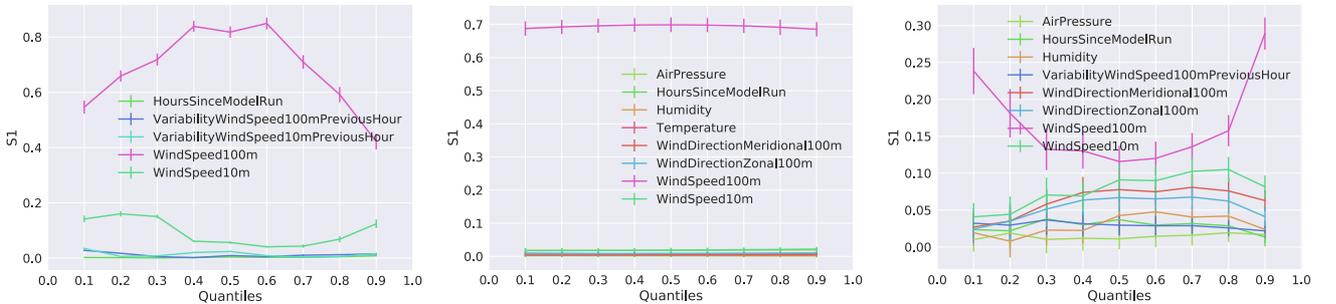

**Figure 1:** The figure shows three representative examples of the individual influences of features on the output of the respective model. For each forecasted quantile Eq. 1 is applied in the following three scenarios, from left to right: (a) GBRT offshore scenario, (b) MCQRNN forest, and (c) SVR flatland.

The individual values (for each PF model) of the sensitivity indexes for different quantiles suggest that it is beneficial to select individual models for different purposes. GBRT, e.g., could be used in a non-complex terrain, to derive a simple model only depending on WS10m and WS100m.

Finally, it seems beneficial to select the model where the influences and the relationship of influences in different quantiles fit our needs. MQRNN, e.g., could be used for probabilistic simulations of the electrical grid for load flow calculations of future energy systems. MQRNN would limit the number of potential influences for different quantiles to provide the simplest possible model for the simulation.

## V. CONCLUSION AND FUTURE WORK

In this article, we proposed a simple method to apply SA to ECDF, predicted by PF models, by applying SA to each quantile estimate individually.

By applying the SA to three PF methods and for data of 28 wind parks within three use cases we moved from a black-box to a gray-box probabilistic forecasting model.

We show that influences on quantile estimates and the relationship of those influences for different quantiles depends on the underlying PF model. Further on, we show that influences are more similar for equal PF models in different use cases than they are for different PF models applied to the same use case. The similarity is either related to the pre-selection of input features (specific to the use case) or the internal behavior between input and output modeled by the PF technique. The observed similarity could, e.g., analyzed further by using the same input features for all use cases in the future.

In our future work, we also aim to investigate the differences between model types further. Primarily, we are interested in analysis distinct types of so-called multi-task (MQRNN) approaches such as hard and soft parameter sharing.
## ACKNOWLEDGMENT

This work was supported within the project Prophesy (0324104A) funded by BMWi (Deutsches Bundesministerium für Wirtschaft und Energie / German Federal Ministry for Economic Affairs and Energy).